\documentclass[letterpaper, 10 pt, conference]{ieeeconf}  

\IEEEoverridecommandlockouts                              

\usepackage{float}

\usepackage{graphicx}
\usepackage{subfigure}
\usepackage[colorlinks,linkcolor=magenta]{hyperref}
\usepackage{amsmath} 
\usepackage{amssymb}  
\usepackage{booktabs}  
\usepackage{multirow}
\usepackage{makecell}

\title{\LARGE \bf
Towards Open-set Gesture Recognition via Feature Activation Enhancement and Orthogonal Prototype Learning
}

\author{Chen Liu$^{\dag}$, Can Han$^{\dag}$, Chengfeng Zhou, Crystal Cai, Suncheng Xiang, Hualiang Ni and Dahong Qian$^{*}$
\thanks{$\dag$These authors contributed to the work equally and should be regarded as co-first authors.}
\thanks{${*}$Corresponding author.}
\thanks{$^{1}$This work was partially supported by OYMotion Technologies.}
\thanks{$^{2}$Chen Liu, Can Han, Chengfeng Zhou, Crystal Cai, Suncheng Xiang and Dahong Qian are affiliated with the School of Biomedical Engineering, Shanghai Jiao Tong University, Shanghai 200240, China.
        (email address: lchen1206@sjtu.edu.cn; hancan@sjtu.edu.cn; chengfengzhou@sjtu.edu.cn; crystal.cai@sjtu.edu.cn; xiangsuncheng17@sjtu.edu.cn; dahong.qian@sjtu.edu.cn.)}
\thanks{$^{3}$Hualiang Ni is affiliated with OYMotion Technologies, Shanghai 200240, China.
        (email address: hni@oymotion.com.)}%
}

\begin{document}

\maketitle
\thispagestyle{empty}
\pagestyle{empty}

\begin{abstract}

Gesture recognition is a foundational task in human-machine interaction (HMI). While there has been significant progress in gesture recognition based on surface electromyography (sEMG), accurate recognition of predefined gestures only within a closed set is still inadequate in practice. It is essential to effectively discern and reject unknown gestures of disinterest in a robust system. Numerous methods based on prototype learning (PL) have been proposed to tackle this open set recognition (OSR) problem. However, they do not fully explore the inherent distinctions between known and unknown classes.
In this paper, we propose a more effective PL method leveraging two novel and inherent distinctions, feature activation level and projection inconsistency. Specifically, the Feature Activation Enhancement Mechanism (FAEM) widens the gap in feature activation values between known and unknown classes.
Furthermore, we introduce Orthogonal Prototype Learning (OPL) to construct multiple perspectives.
OPL acts to project a sample from orthogonal directions to maximize the distinction between its two projections, where unknown samples will be projected near the clusters of different known classes while known samples still maintain intra-class similarity.
Our proposed method simultaneously achieves accurate closed-set classification for predefined gestures and effective rejection for unknown gestures.
Extensive experiments demonstrate its efficacy and superiority in open-set gesture recognition based on sEMG.

\end{abstract}

\section{INTRODUCTION}

In recent years, the study of human gestures based on cameras and sensors has gained increasing significance in human-machine interaction (HMI)~\cite{Sun2022Metaphoraction, Carfi2023GestureHMI,han2023mtdl}.
In this interaction paradigm, gesture recognition serves as a foundational task and has been extensively applied across diverse domains, such as rehabilitation robotics~\cite{Copaci2022sEMGBasedGC,Liu2021AFC}, mobile interaction~\cite{Xiong2021Review, lin2022interactive}, virtual reality~\cite{Gao2022KeyTO} and exoskeleton control~\cite{kim2022semg}.

Surface electromyography (sEMG) signals can reflect muscle activity related to motor function and have been a popular method to monitor human motion due to its non-invasiveness along with its convenience of data acquisition.
Recently, the development of sEMG-based gesture recognition systems~\cite{Tsinganos2019ImprovedGR,karnam2022emghandnet,mendes2022surface} has been remarkable.
However, most of them are confined to closed-set scenarios, where the training and test sets share an identical label space~\cite{xiang2020unsupervised,xiang2023less}.
These closed-set systems lack robustness and reliability in the dynamic and ever-changing real world, which causes them to mistake novel gestures or unintentional motions as known ones and generate false interaction signals.
Therefore, a robust gesture recognition system, one which can correctly classify predefined known gestures while identifying unknown gestures in real-world scenarios, is in high-demand.

\begin{figure}[!t]
\centering
\vspace{0.3cm}
\includegraphics[width=1.0\linewidth]{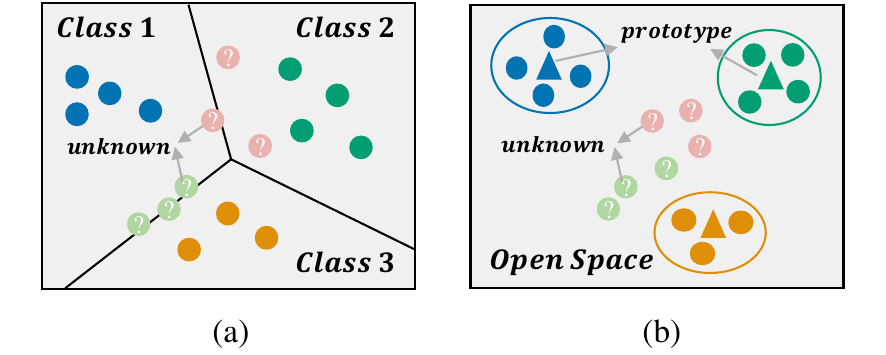}
\vspace{-0.8cm}
\caption{\textbf{An illustration of the advantages of a compact feature space.} (a) Softmax-based classification models partition all space for known classes. (b) Prototype learning constructs a compact feature space to build closed classification boundaries and to leave open space for unknown classes.}
\label{fig1}
\end{figure}

Scheirer~\cite{Scheirer2013Open} first described the above demand as open set recognition (OSR), whose test set contains unknown classes that are not included in the training set.
Scheirer proposed that the key for OSR is to make the model aware of the existence of unknown samples and reserve open space for them, where open space is the space far from known samples~\cite{Scheirer2013Open}.
Softmax-based classification models do not leave open space for unknown samples, since they divide the feature space into several semi-open subspaces as shown in Fig.~\ref{fig1}(a).
To this end, methods based on prototype learning (PL)~\cite{Yang2020ConvolutionalPN,Chen2022ARPL,Xia2023AMPF,Cevikalp2023From} are popular in OSR because they aim to create a compact feature space for known classes which aids in establishing a closed classification boundary and leaves open space for unknown samples, as shown in Fig.~\ref{fig1}(b).
These methods primarily rely on the differences in feature distribution between known and unknown classes to reject the latter, whereby unknown samples tend to cluster around the ``center" of the embedding space while known samples tend to distribute towards the periphery~\cite{Chen2022ARPL,Xia2023AMPF}.
Although the aforementioned methods based on PL have achieved promising performance, they do not fully explore the inherent distinctions between known and unknown classes.

To tackle open-set gesture recognition, we propose a more effective PL method leveraging two novel and inherent distinctions, feature activation level and projection inconsistency.
In line with the familiarity hypothesis~\cite{dietterich2022familiarity}, unknown samples with few familiar features result in low model responses, which is evident in the feature activation level.
Considering this property, we introduce a novel Feature Activation Enhancement Mechanism (FAEM) to strengthen the distinction between feature activation levels of known and unknown classes.

\begin{figure*}[thpb]

\centering
\includegraphics[scale=0.6]{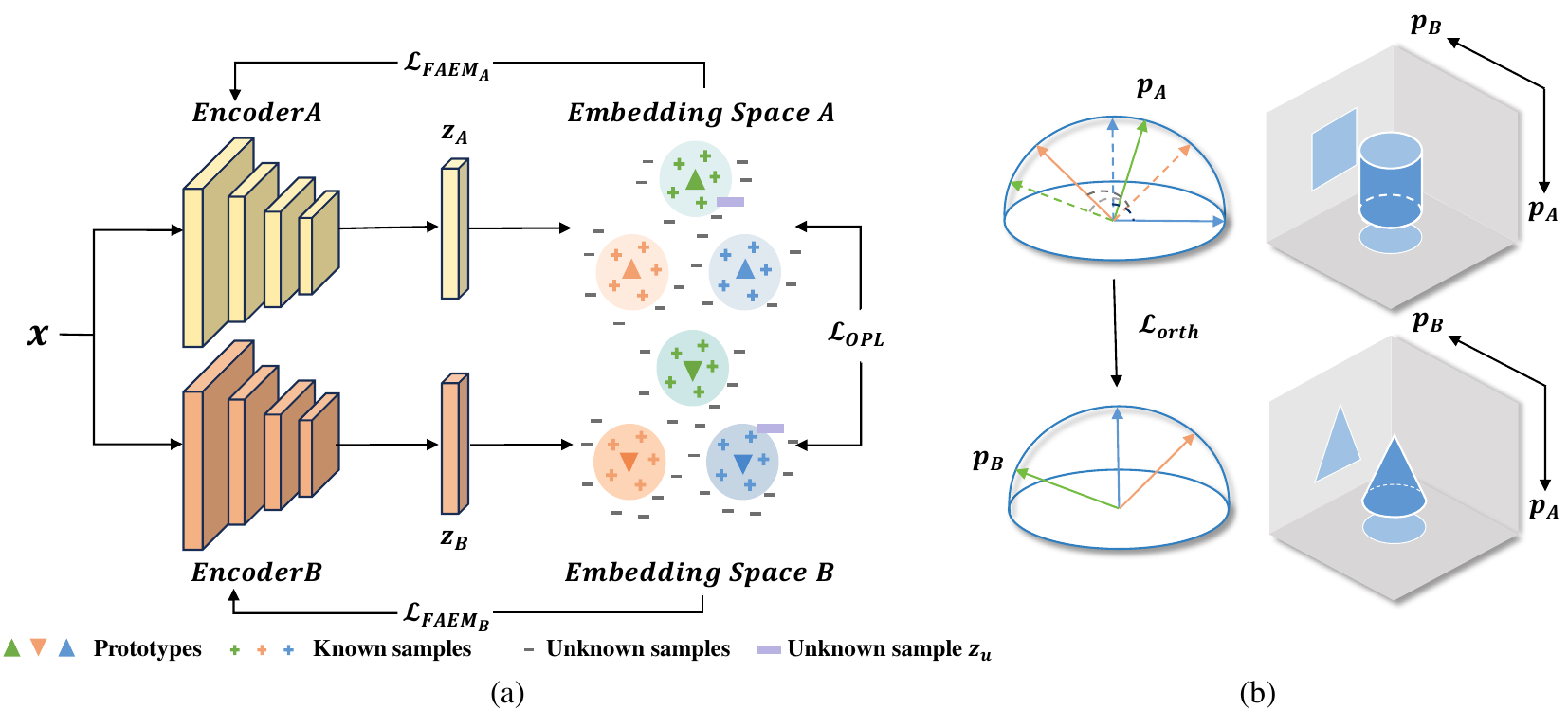}
\vspace{-0.4cm}
\caption{\textbf{An illustration of our proposed framework combining FAEM and OPL.} (a) Our framework builds two branches, each of which contains an encoder and a set of learnable prototypes. $\mathcal{L}_{FAEM}$ is applied to each branch individually while $\mathcal{L}_{OPL}$ simultaneously acts on both. Based on the PL framework, unknown samples like $\mathbf{z}_{u}$ represented by purple are projected near the clusters of different known classes due to their projection inconsistency, while known samples maintain their intra-class similarity across two branches. (b) A toy example for illustrating the projection inconsistency. We maximize the projection inconsistency by $\mathcal{L}_{orth}$ which enforces the same-class prototypes $\mathbf{p}_A$ and $\mathbf{p}_B$ of the two branches to be orthogonal.}
\label{fig2}
\end{figure*}

However, certain unknown samples with a greater number of familiar features will exhibit pseudo-similarity to known classes, resulting in a falsely high feature activation level.
Incorporating the projection inconsistency across multiple perspectives is advantageous for further enhancing the differentiation between known and unknown classes. 
Specifically, we could represent a known sample as a cone and an unknown sample as a cylinder.
Imagine projecting the cone and cylinder as shown in Fig.~\ref{fig2}(b): the projection of the cylinder appears identical to the cone when viewed from the top, but different from another angle.
Nonetheless, the projection of the cone retains its resemblance to other cones, regardless of the projection direction.
Similarly, a known sample will be consistent with other known samples of the same class from any projection direction, constituting intra-class similarity.
However, an unknown sample may resemble a known class from one perspective but another known class from a different perspective, leading to its projection inconsistency, as shown in Fig.~\ref{fig2}(a).
A complement orthogonal perspective will ensure maximum projection inconsistency and avoid multi-perspective inefficiencies.
To this end, we construct an Orthogonal Prototype Learning (OPL) framework which projects samples from orthogonal directions to further identify unknown samples by their projection inconsistencies.
Moreover, background samples are introduced as prior knowledge about the unknown to reduce feature activation values of unknown samples and amplify their projection inconsistencies. 

To our knowledge, this is the first attempt to integrate open set scenarios into sEMG-based gesture recognition tasks. Our contributions are as follows:
\begin{itemize}
    \item [1)] In this paper, we propose a new method, FAEM, to tackle OSR by significantly widening the gap between feature activation levels of known and unknown classes.
    \item [2)] Furthermore, we propose a novel framework, OPL, which effectively distinguishes between known and unknown classes by leveraging the projection inconsistency of the unknown.
    \item [3)]  Comprehensive experiments on benchmarks demonstrate that our approach simultaneously achieves the maintenance of classification accuracy for known gestures and effective rejection for unknown gestures, outperforming previous approaches with a clear margin.
\end{itemize}

\section{Related Work}

\subsection{sEMG-based Gesture Recognition}

The emergence of deep learning has freed sEMG-based gesture recognition from the constraints of manual feature extraction~\cite{Xiong2021Review}, facilitating a better understanding of human gesture. Various deep learning architectures have been widely employed for this task. Park et al.~\cite{Park2016MovementID} pioneered the application of Convolutional Neural Network (CNN) models to classify the Ninapro DB2 dataset~\cite{DB2}. Furthermore, more complex CNN models and Recurrent Neural Network (RNN) models have showcased their superiority to fine gesture classification~\cite{Tsinganos2019ImprovedGR,karnam2022emghandnet,mendes2022surface}. EMGHandNet~\cite{karnam2022emghandnet} proposed a hybrid CNN and Bi-LSTM framework to capture both the inter-channel and temporal features of sEMG, achieving classification accuracy of $92.01\%$ on $49$ gestures. However, these extraordinary performances of classic closed-set systems are illusory, since their applications are limited when it comes to real and open world. Only a few studies~\cite{Wu2022RejectingNM,Wu2022UnknownMR} have focused on rejecting novel gestures, but they overlook the importance of maintaining closed-set classification accuracy.

\subsection{Open Set Recognition}

Open set recognition seeks to generalize the recognition tasks from a closed-world assumption to an open set. Scheirer~\cite{Scheirer2013Open} first formalized OSR as a constrained minimization problem and proposed the ``1-vs-Set Machine” method. A critical aspect of OSR is to explore the distinctions between known and unknown classes. Previous studies have found several distinctions in prediction probability, reconstruction error, uncertainty and distance. Specifically, methods based on prediction probability~\cite{Bendale2016Openmax,Zhou2021Palceholder} reject unknown samples with low probability. Methods based on reconstruction error~\cite{Perera2020Geneartive,huang2022class} assume that generative models trained on known classes cannot reconstruct unknown samples effectively. Uncertainty-based methods~\cite{Bao2021EvidentialDL} directly model the uncertainty of samples and identify those with high uncertainty as unknown classes. A recently popular trend for OSR is employing PL to learn a compact feature space and to establish a clear distance boundary between the known and unknown~\cite{Yang2020ConvolutionalPN,Chen2022ARPL,Xia2023AMPF,Cevikalp2023From}. Numerous researchers believe that modeling only known classes is insufficient and suggest that incorporating prior knowledge about unknown classes is necessary. Some approaches attempt to generate fake data~\cite{moon2022difficulty}, counterfactual images~\cite{Neal2018OSCI} or confused samples~\cite{Chen2022ARPL,Xia2023AMPF}. Others~\cite{Dhamija2018Reducing,cho2022towards} introduce background classes or known unknown classes (KUC) to represent the unknown.

\subsection{Prototype Learning}

Yang et al.~\cite{Yang2020ConvolutionalPN} first introduced prototype learning into convolution networks to tackle OSR.
However, the concept of prototypes can be traced back earliest to Learning Vector Quantization (LVQ)~\cite{Liu2001EvaluationOP}, where they are used to represent the average or center of a class.
These methods encourage samples to be close to their corresponding prototypes and far apart from other prototypes, thereby establishing a compact feature space and a closed boundary. Different methods show variations in the ways of prototype initialization and optimization. Recent methods~\cite{Wang2019PushFC,Shen2023EquiangularBV} utilize orthogonal constraints on prototypes to widen inter-class separability. However, these methods only utilize a single perspective while the benefits of bringing multiple perspectives are still lacking in exploration.



\section{Methodology}

\subsection{Problem Definition}

 Considering the sEMG-based gesture recognition in real-world scenarios, we assume that $ \mathcal{Y} \subset \mathbb{N} $ is the infinite label space of all possible gesture classes. Assume that $\mathcal{C} = \{1,\ldots,N\} \subset \mathcal{Y}$ represents $N$ known classes of interest. The set $\mathcal{U} = \mathcal{Y} \backslash \mathcal{C}$ represents all unknown classes that need to be rejected. Since $ \mathcal{Y}$ is infinite and $\mathcal{C}$ is finite, $\mathcal{U}$  is also infinite. The objective of open set recognition is to find a measurable recognition function $f^*\subset \mathbb{H}$ which minimizes both the empirical classification risk on known samples and the open space risk on unknown samples:
\begin{equation}
f^*={ \underset{f}{\arg\min}} \{R_\epsilon(f,D_c) + \lambda R_O(f, D_u)\},
\label{eq1}
\end{equation}
where $D_c$ and $D_u$ represent samples belonging to known and unknown classes, respectively.



\subsection{Methodology Overview}
Our proposed framework is illustrated in Fig.~\ref{fig2}(a). To minimize both the empirical classification risk and open space risk simultaneously, we present FAEM and OPL to leverage the feature activation level and projection inconsistency distinctions. Specifically, our framework contains two branches for extracting features from different perspectives. Each branch contains an encoder and $N$ learnable prototypes of known classes. The FAEM is applied to each branch to amplify the disparity between feature activation levels of known and unknown classes. OPL operates concurrently on both branches to extract features from orthogonal projection directions, thereby further revealing distinctions between known and unknown classes, as illustrated in Fig.~\ref{fig2}(b). The encoders can be arbitrary network architectures for extracting features from sEMG signals, both having identical structures but different weights.

\subsection{Feature Activation Enhancement Mechanism}
\label{sec3.2}
Following the PL method~\cite{Yang2020ConvolutionalPN,Chen2022ARPL}, we define the embedding feature of sample $\mathbf{x}_i$ as $\mathbf{z}_i = f(\mathbf{x}_i) \in \mathbb{R}^d$. $N$ known classes are each assigned a learnable prototype $\mathbf{p}_k \in \mathbb{R}^d$, where $1 \leq k \leq N$. The probability of $\mathbf{x}_i$ belonging to class $k$ is based on the distance $d(f(\mathbf{x}_i),\mathbf{p}_k)$:
\begin{equation}
p(y_i=k | \mathbf{x}_i, f, \mathbf{p}) = \frac{e^{-d(f(\mathbf{x}_i),\mathbf{p}_k)}}{\sum_{j=1}^N e^{-d(f(\mathbf{x}_i), \mathbf{p}_j)}}.
\label{eq2}
\end{equation}

To narrow the distance between samples and their corresponding prototypes while pushing them away from other prototypes, the DCE loss function~\cite{Yang2020ConvolutionalPN} is utilized and described as follows:
\begin{equation}
\mathcal{L}_\epsilon = -\log{(p(y=k|\mathbf{x},f,\mathbf{p}))} .
\label{eq3}
\end{equation}

We use the dot product to measure the generalized distance between the samples and prototypes,
\begin{equation}
d(\mathbf{z}_i,\mathbf{p}_k) = -\mathbf{z}_i \cdot {\mathbf{p}_k}^T .
\label{eq4}
\end{equation}

The loss function $\mathcal{L}_\epsilon$ also optimizes prototypes to enhance their separation, which increases the inter-class separability and thereby reduces the empirical classification risk. On the other hand, this distance setting tends to make prototypes $\mathbf{p}_k$ converge to a higher activation level, leading to a higher activation level of known classes as well. But for unknown samples, models trained on known samples struggle to extract features relevant to any known class from them, leading to their low feature activation levels~\cite{dietterich2022familiarity}. Therefore, we develop the FAEM to leverage and emphasize this property, expecting the feature activation level of known samples to trend towards higher values and that of unknown ones towards lower values.

In order to further increase the feature activation level of known classes, we align the features $\mathbf{z}_i$ of known classes with their corresponding prototypes $\mathbf{p}_k$, which encourages their features to be close to the corresponding prototypes in activation values. To prevent training oscillation, a smoothing loss function is designed, as shown below:
\begin{equation}
\mathcal{L}_{F} = \frac{1}{M}\sum_{i=1}^M \mathcal{L}_n(f(\mathbf{x}_i)-\mathbf{p}_k),
\label{eq5}
\end{equation}
in which
\begin{equation}
\mathcal{L}_n(\mathbf{u}) =
\begin{cases}
\frac{1}{2} \Vert \mathbf{u} \Vert_2 &\quad \Vert \mathbf{u} \Vert_1 < 1\\
\Vert \mathbf{u} \Vert_1 -\frac{1}{2} &\quad \Vert \mathbf{u} \Vert_1 \geq 1 \\
\end{cases},
\label{eq6}
\end{equation}
where $M$ represents the number of known samples.

While the model ensures high feature activation level for known classes, it lacks constraints on unknown ones. Incorporating prior knowledge about unknown classes will further reduce open space risk~\cite{Cevikalp2023From,Dhamija2018Reducing,cho2022towards}. Following this strategy, we introduce background samples $\mathbf{x}_{bi}$ during training and align them with the center point $\mathbf{p}_c$. This alignment is consistent with the distribution of unknown classes in the embedding space, where they are closer to the center point in distance. The center point $\mathbf{p}_c$ is calculated as the mean of all prototypes, and its feature activation value remains low during optimization, ensuring that unknown samples have a lower feature activation level compared with the known. The loss function is designed as follows:
\begin{equation}
\mathcal{L}_{Fb} = \frac{1}{M_b}\sum_{i=1}^{M_b} \mathcal{L}_n(f(\mathbf{x}_{bi})-\mathbf{p}_c),
\label{eq7}
\end{equation}
where $M_b$ represents the number of background samples.

Combining~\eqref{eq3},~\eqref{eq5} and~\eqref{eq7}, the overall loss function of the FAEM is expressed as follows:
\begin{equation}
\mathcal{L}_{FAEM} = \mathcal{L}_\epsilon+\lambda \mathcal{L}_{F} + \gamma \mathcal{L}_{Fb},
\label{eq8}
\end{equation}
where $\lambda$ and $\gamma$ are the weights of each module.

\subsection{Orthogonal Prototype Learning}

While the FAEM enhances the distinction in feature activation level, the model might unintentionally capture certain familiar features from unknown samples, making these samples exhibit pseudo-similarity to a certain known class. Eliminating this pseudo-similarity solely based on a single perspective presents a significant challenge. However, it is also difficult for unknown samples to maintain this pseudo-similarity with the same known class across multiple perspectives, since unknown samples do not entirely possess the features of known classes.

Therefore, we propose OPL to reduce the pseudo-similarity by extending PL to multiple projections. Specifically, OPL simulates multi-projection by employing encoders with different weights, since encoders generally act to project samples from high-dimensional sample space to low-dimensional embedding space. Orthogonal projections can maximize differences between perspectives, thereby reducing the overlap between features learned from the two branches. This contributes to separately projecting unknown samples near the clusters of different known classes within the two branches while known samples will still maintain intra-class similarity, as shown in Fig.~\ref{fig2}(a). Combined with PL, it is easy to achieve orthogonal projections by forcing the prototypes of the two branches $A$ and $B$ to be orthogonal (Fig.~\ref{fig2}(b)), since the essence of PL is to bring samples and their corresponding prototypes closer together. The loss function is designed as follows:
\begin{equation}
\mathcal{L}_{orth} = \frac{1}{N} \sum_{k=1}^N ({\mathbf{p}_{Ak} \cdot {\mathbf{p}_{Bk}}^T})^2,
\label{eq9}
\end{equation}
where $N$ represents the number of known classes.

To further enhance the projection inconsistency of unknown classes in OPL, we introduce a penalty mechanism leveraging background class information. Specifically, if a background sample is close to the same-class prototypes in the two branches, we will push it away from the closest prototype in one of the branches. The closeness degree is based on generalized distance as~\eqref{eq4}. This further encourages a background sample to be classified into different known classes in two projections, thereby enhancing projection inconsistency. The penalty loss function is described as follows:
\begin{equation}
\mathcal{L}_{Pb} = \frac{1}{M_{pb}}\sum_{i=1}^{M_{pb}} \mathbf{z}_{bi} \cdot {\mathbf{p}_k}^T,
\label{eq10}
\end{equation}
where $M_{pb}$ is the number of background samples close to the same-class prototypes across two branches.

Combined with the FAEM, each branch adopts the loss function as~\eqref{eq8}. The following loss function applies to both branches simultaneously:
\begin{equation}
\mathcal{L}_{OPL} = \alpha \mathcal{L}_{orth} + \beta \mathcal{L}_{Pb},
\label{eq11}
\end{equation}
where $\alpha$ and $\beta$ are the hyperparameters that control the weights.

\subsection{Unknown Detection}
Most previous PL-based approaches use similarity based on distance between sample features and prototypes to detect unknown samples. Specifically, similarity between prototypes $\mathbf{p}_k$ and features $\mathbf{z}_i$ for each branch is defined as follows:
\begin{equation}
\textrm{Sim}(\mathbf{z}_i,\mathbf{p}_k) = \mathbf{z}_i \cdot {\mathbf{p}_{k}}^T.
\label{eq12}
\end{equation}

The proposed FAEM allows us to differentiate between known and unknown classes by combining feature activation level. Therefore, at each branch we incorporate the similarity and feature activation level to facilitate the detection of unknown classes. Specifically, the known confidence $\textrm{Score}(\mathbf{z}_i,\mathbf{p}_k)$ is defined to describe the confidence that the sample $\mathbf{x}_i$ belongs to the class $k$:
\begin{equation}
\textrm{Score}(\mathbf{z}_i,\mathbf{p}_k) = \textrm{Sim}(\mathbf{z}_i,\mathbf{p}_{k}) * \Vert \mathbf{z}_i \Vert_1.
\label{eq13}
\end{equation}

More importantly, utilizing the projection inconsistency across two branches of OPL can further aid in detecting unknown samples. So, we combine the known confidence of the two branches:
\begin{equation}
C_k = {\textrm{Score}(\mathbf{z}_{Ai},\mathbf{p}_{Ak})} + {\textrm{Score}(\mathbf{z}_{Bi},\mathbf{p}_{Bk})}.
\label{eq14}
\end{equation}

In order to classify sample $\mathbf{x}_i$ as a certain known class $k$ or reject it as unknown classes, we further obtain $C_{max}$:
\begin{equation}
C_{max} =  C_{k^*},
\label{eq13}
\end{equation}
where
\begin{equation}
k^* = \underset{1 \leq k \leq N}{\arg\max} \ C_k.
\label{eq13}
\end{equation}

In summary, we comprehensively consider the distinctions between known and unknown classes based on similarity along with feature activation level and projection inconsistency, thereby effectively increasing rejection of the unknown. Specifically, known samples will exhibit high feature level and be close to the same-class prototypes in both branches resulting in high $C_{max}$, while unknown samples will obtain low $C_{max}$ due to their low feature activation level and projection inconsistency. A pre-determined threshold can be applied to $C_{max}$ to reject the unknown. Samples with $C_{max}$ greater than the threshold value will be classified as class $k^*$.

\section{Results And Discussion}
\subsection{Experimental Setting}
To comprehensively extract sEMG features, we employ a network that combines a ResNet variant~\cite{He2016Resnet}, an LSTM~\cite{Ho1997LSTM} and an SKAttention~\cite{Li2019SK}.
During experiments, we used the SGD optimizer with a learning rate of $0.01$. The batch size is set to $256$.
The hyperparameters $\lambda$, $\gamma$ in~\eqref{eq8}, $\alpha$ and $\beta$ in~\eqref{eq11} are empirically set to $1$, $1$, $0.1$ and $0.01$, respectively, while the feature dimension of embedding space is $128$.
Prototypes are randomly initialized by the standard normal distribution.

\subsection{Datasets and Evaluation Metrics}
Four public benchmark datasets~\cite{OrtizCataln2013BioPatRecAM,DB2,DB4,DB7} are applied to validate the proposed approach, as shown in Table~\ref{tab1}. Following the protocol of~\cite{Scheirer2013Open}, we randomly selected $10$ known classes from BioPat DB2 and $15$ known classes from Ninapro DB2, Ninapro DB4 and Ninapro DB7, while treating the remaining classes as unknown.
Within the set of unknown classes, one is randomly selected as the background class.

The evaluation metrics used for OSR are derived from~\cite{Chen2022ARPL,Xia2023AMPF}, including the area under the receiver operating characteristic (AUROC) for measuring the ability to distinguish between known and unknown classes, open set classification rate (OSCR) based on classification accuracy (CCR) and false positive rate (FPR) to comprehensively evaluate empirical classification risk and open space risk, and closed-set accuracy (ACC) for assessing known classes classification performance.


\begin{table}[!t]
    \centering
    \caption{Characteristics and setup of four public datasets.}
    \small
    \setlength{\tabcolsep}{0.95mm}{
        \begin{tabular}{lccccc}
        \toprule
        Dataset & Subject & Channel & Gesture & Trial \\
        \midrule
        BioPat DB2 & 17 & 8 & 26 & 3  \\
        Ninapro DB2 & 40 & 12 & 49 & 6   \\
        Ninapro DB4 & 10 & 12 & 52 & 6   \\
        Ninapro DB7 & 20 & 12 & 40 & 6  \\
        \bottomrule
        \end{tabular}}
        \label{tab1}
\end{table}

\begin{table*}[!t]
    \centering
    \caption{Performance comparison (\%) with SOTAs in terms of AUROC, OSCR and closed-set ACC on four public datasets. Results are averaged among five randomized trials. Best performances are highlighted in bold, while the second best with underlined.}
    \small
    \setlength{\tabcolsep}{0.95mm}{
        \begin{tabular}{lcccccccccccc}
        \toprule
        \multirow{2}[4]{*}{\textbf{Methods}} & \multicolumn{3}{c}{\textbf{BioPat DB2}}     & \multicolumn{3}{c}{\textbf{Ninapro DB2}} & \multicolumn{3}{c}{\textbf{Ninapro DB4}} & \multicolumn{3}{c}{\textbf{Ninapro DB7}}\\
        \cmidrule(lr){2-4} \cmidrule(lr){5-7} \cmidrule(lr){8-10} \cmidrule(lr){11-13}
        & \textbf{AUROC}   & \textbf{OSCR}  & \textbf{ACC} & \textbf{AUROC}   &  \textbf{OSCR}  & \textbf{ACC}  & \textbf{AUROC}   & \textbf{OSCR} & \textbf{ACC}  & \textbf{AUROC}   &  \textbf{OSCR} & \textbf{ACC}\\
        \midrule
        Softmax & 85.15 & 85.09 & 99.71 & 81.71 & 80.52 &96.48 & 83.86 & 82.77 & 96.81 & 81.75 & 80.73 & 96.98\\
        OpenMax~\cite{Bendale2016Openmax} & 86.36 & 85.88 & 96.37 & 83.91 & 83.13 & 97.67 & 84.61 & 83.44 & 96.14 & 81.55 & 77.88 & 93.70\\
        ARPL~\cite{Chen2022ARPL}    & 86.81 & 86.73 & 99.72 & 87.10 & 86.58 & 98.72 & 86.49 & 85.86 & 98.28 & 87.10 & 86.68 & 98.95\\
        ARPL+CS~\cite{Chen2022ARPL} & 85.69 & 85.60 & 99.70 & 88.61 & 88.17 & \underline{98.89} & 88.05 & 87.45 & 98.45 & 88.33 & 87.89 & 98.95 \\
        AKPF~\cite{Xia2023AMPF}    & 87.19 & 87.12 & \underline{99.74} & 87.16 & 86.65 & 98.76 & 86.54 & 85.87 & 98.22 & 87.41  & 86.95 & 98.85 \\
        AKPF+CS~\cite{Xia2023AMPF} & \underline{88.38} & \underline{88.31} & 99.70 & 86.42 & 85.62 & 98.06 & 87.46 & 86.42 & 97.54 & 87.52 & 86.95 & 98.59 \\
        Objectosphere~\cite{Dhamija2018Reducing}  & 87.38 & 87.26 & 98.64 & 87.65 & 86.98 & 98.49 & 86.88 & 86.06 & 97.93 & 87.76 & 87.16 & 98.62 \\
        DIAS~\cite{moon2022difficulty}  & 86.06 & 85.98 & 99.69 & \underline{89.04} & \underline{88.50} & 98.46 & \underline{90.53} & \underline{90.06} & \underline{98.61} & \underline{90.54} & \underline{90.28} & \textbf{99.24} \\
        \midrule
        FAEM+OPL(ours) & \textbf{91.34} & \textbf{91.31} & \textbf{99.85} & \textbf{92.48} & \textbf{92.07} & \textbf{98.97} & \textbf{93.96} & \textbf{93.71} & \textbf{99.34} & \textbf{92.32} & \textbf{92.00} & \underline{99.19} \\
        \bottomrule
        \end{tabular}}
        \label{tab2}
        \vspace{-0.3cm}
\end{table*}

\begin{table}
    \centering
    \caption{Ablations of each module in terms of AUROC (\%) on four public datasets. MP represents multiple projections. Results are averaged among five randomized trials.}
    \small
    \setlength{\tabcolsep}{0.95mm}{
        \begin{tabular}{ccccccccc}
        \toprule
         \multicolumn{5}{c}{\textbf{Methods}} & \multirow{2}[4]{*}{\textbf{\makecell[c]{BioPat \\ DB2}}} & \multirow{2}[4]{*}{\textbf{\makecell[c]{Ninapro \\ DB2}}} & \multirow{2}[4]{*}{\textbf{\makecell[c]{Ninapro \\ DB4}}} & \multirow{2}[4]{*}{\textbf{\makecell[c]{Ninapro \\DB7}}} \\
         \cmidrule{1-5} MP & $\mathcal{L}_{F}$  & $\mathcal{L}_{Fb}$  & $\mathcal{L}_{orth}$ & $\mathcal{L}_{Pb}$ \\
        \midrule
          $\times$ &$\times$ &$\times$ & $\times$ & $\times$& 87.18  & 87.28 & 85.81 & 87.04 \\
          $\times$&\checkmark  & $\times$  &  $\times$ & $\times$ & 88.17 & 88.14 & 89.67 & 87.44  \\
          $\times$&\checkmark  & \checkmark  & $\times$  & $\times$    & 88.59 & 88.38 & 90.45  & 88.78 \\
         \checkmark &\checkmark  & \checkmark  & $\times$ & $\times$  & 90.34 & 92.14 & 93.65 & 91.73  \\
         \checkmark &\checkmark  & \checkmark  & \checkmark  & $\times$ & 91.09 & 92.29 & 93.82 & 92.14  \\
        \midrule
         \checkmark &\checkmark  & \checkmark  & \checkmark  & \checkmark & \textbf{91.34} & \textbf{92.48} & \textbf{93.96} & \textbf{92.32}\\
        \bottomrule
        \end{tabular}}\\
    \label{tab3}

\end{table}

\subsection{Comparison with the State-of-the-arts}

We compared our method against other state-of-the-art open set recognition approaches, including Softmax, OpenMax~\cite{Bendale2016Openmax}, ARPL~\cite{Chen2022ARPL}, AKPF~\cite{Xia2023AMPF}, Objectosphere Loss~\cite{Dhamija2018Reducing} and DIAS~\cite{moon2022difficulty}.
Accordingly, OpenMax calibrates the prediction probability by Weibull distribution.
ARPL and AKPF are methods based on prototype learning.
ARPL+CS and AKPF+CS combine generative models to introduce confused samples.
Objectosphere Loss identifies the unknown based on feature magnitude.
DIAS also introduces a feature generator to produce hard fake instances.
All methods employed the same backbone for a fair comparison.
The results in Table~\ref{tab2} highlight the outstanding performance of our proposed approach for the open-set gesture recognition task. Specifically, our method achieves the best AUROC scores of {\bf91.34\%}, {\bf92.48\%}, {\bf93.96\%} and {\bf92.32\%} across four datasets.
Furthermore, when considering both empirical classification risk and open space risk, our approach also surpasses all SOTA methods, consistently achieving the highest OSCR scores on four datasets.
These results confirm that our method effectively captures the distinctions between known and unknown classes.
Moreover, our method also achieves comparable closed-set accuracy to SOTA methods.
In conclusion, our approach shows the superiority in both closed-set classification and unknown sample rejection.

\subsection{Ablation Study}

As presented in Table~\ref{tab3}, each component within our method has been systematically integrated into the PL framework to verify its necessity.
As demonstrated, the enhancement of feature activation level for known classes yields notable improvements compared to the PL baseline.
The introduction of background samples further strengthens the distinction between feature activation levels of known and unknown classes, resulting in consistent improvements on AUROC.
Based on the FAEM, we construct two branches without any constraints on the prototypes.
This provides an average improvement of {\bf +2.26\%} compared to the FAEM based on a single projection, demonstrating the effectiveness of introducing multiple projections.
Subsequently, applying orthogonal constraints on the two branches can further reveal the projection inconsistency of unknown samples, leading to improved performance.
Similarly, penalizing background samples also exhibits its effectiveness.
Finally, incorporating all above components improves AUROC by at least {\bf +4.16\%} compared to the baseline PL model, which clearly demonstrates that each component contributes to the overall improvement.

\subsection{Discussion}
To provide a comprehensive understanding of our method, we conducted a deep analysis.

\begin{figure}[!t]
    \centering
    \subfigure[]{
        \includegraphics[width=1.58in]{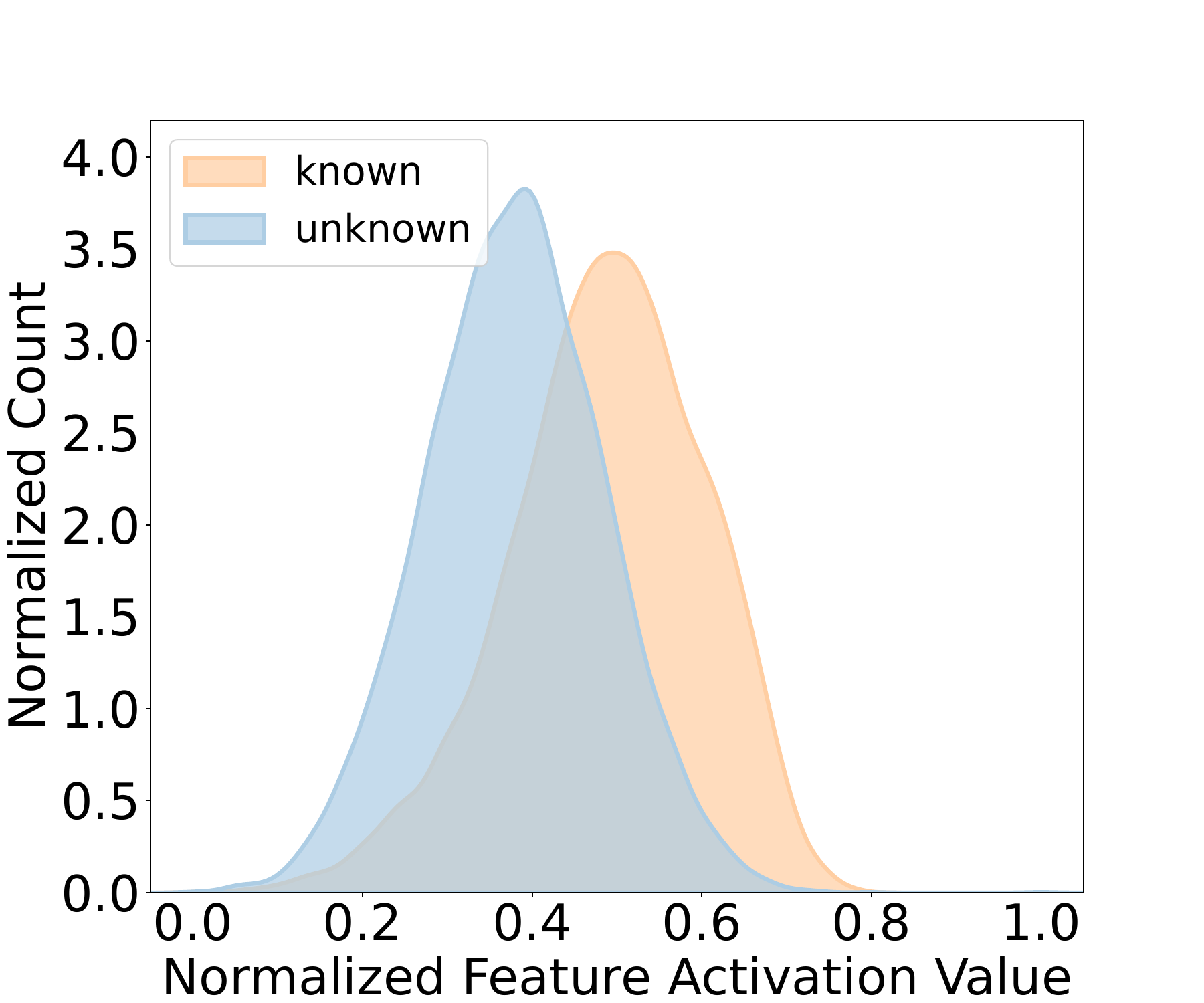}
        \label{label_for_cross_ref_1}
    }
    \subfigure[]{
	\includegraphics[width=1.58in]{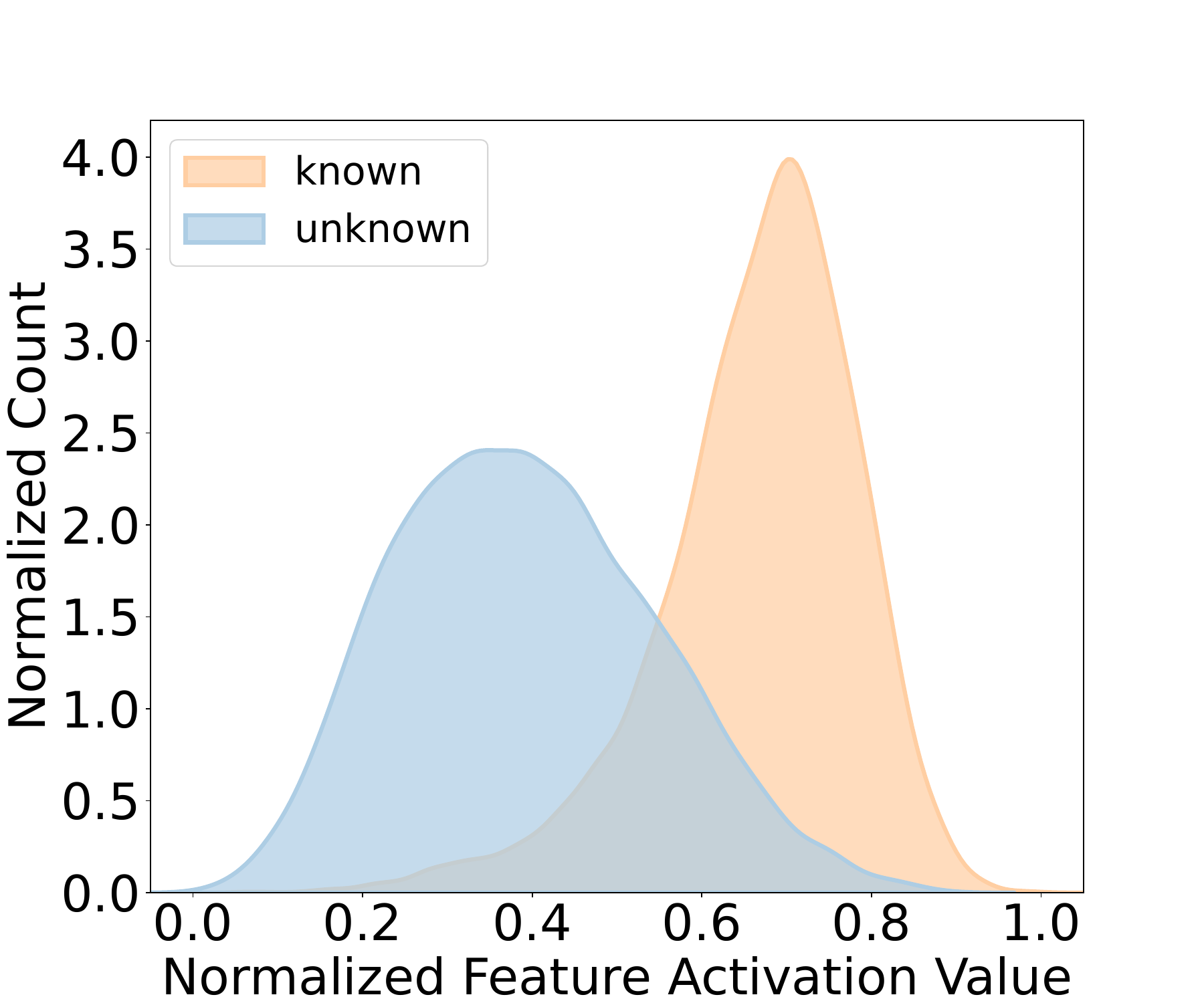}
        \label{label_for_cross_ref_2}
    }
    \vspace{-0.4cm}
    \caption{\textbf{Normalized histograms of feature activation values of DB4 samples.} (a) Baseline PL model. (b) FAEM.}
    \label{fig3}
\end{figure}

\textit{\textbf{Feature Activation Level.}}
Fig.~\ref{fig3} illustrates the normalized histogram of feature activation values for the baseline PL model and our proposed model.
We can observe a notable distinction between feature activation levels of known and unknown samples in the two histograms.
However, the baseline PL model exhibits a substantial overlap in its histogram, thus hampering the discrimination of known and unknown classes.
By introducing FAEM (Fig.~\ref{fig3}(b)), the distinction of feature activation values becomes more pronounced and the extent of overlap is notably diminished. 

\begin{figure}[!t]
    \centering
    \subfigure[]{
        \includegraphics[width=1.58in]{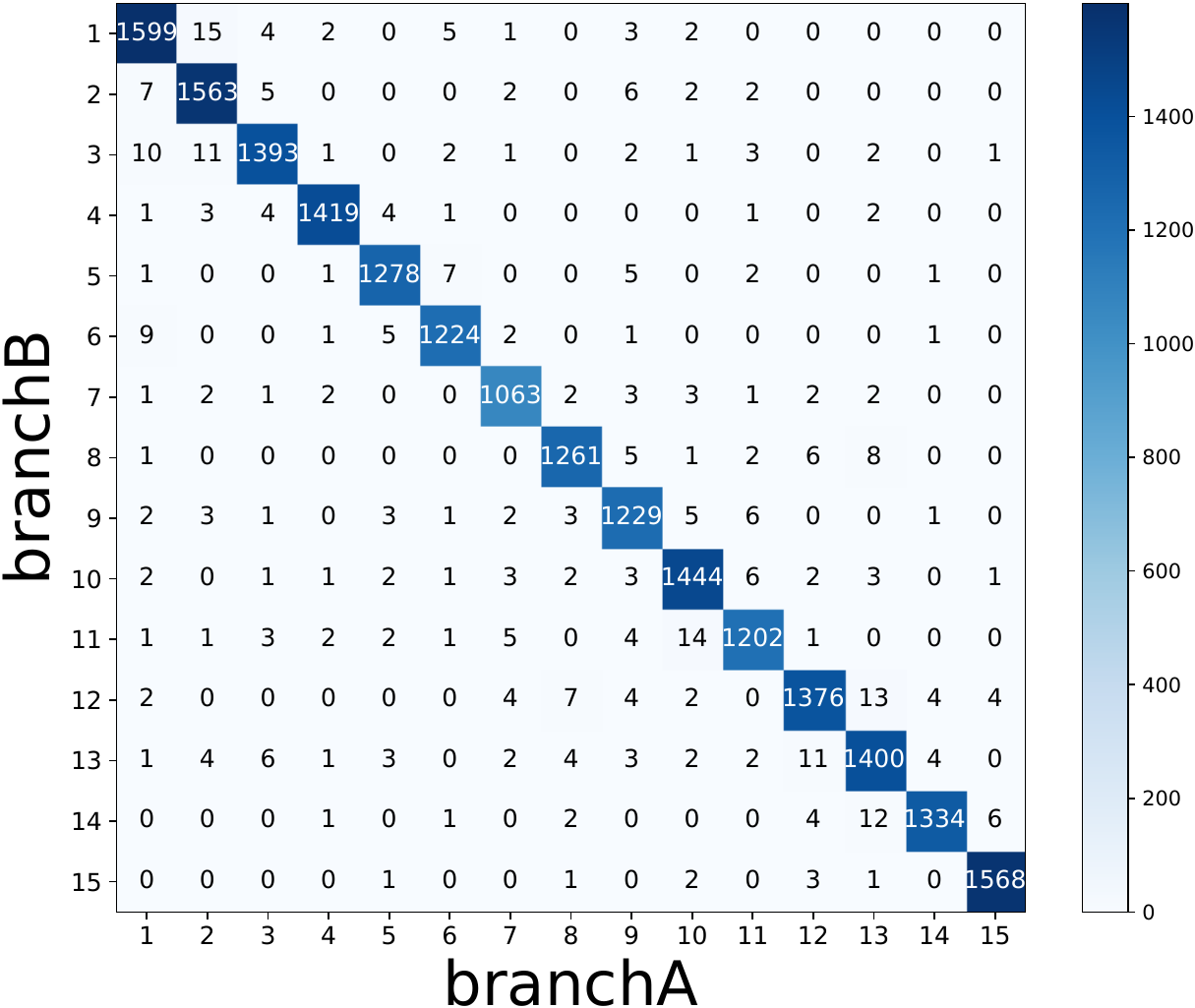}
        \label{label_for_cross_ref_1}
    }
    \subfigure[]{
	\includegraphics[width=1.58in]{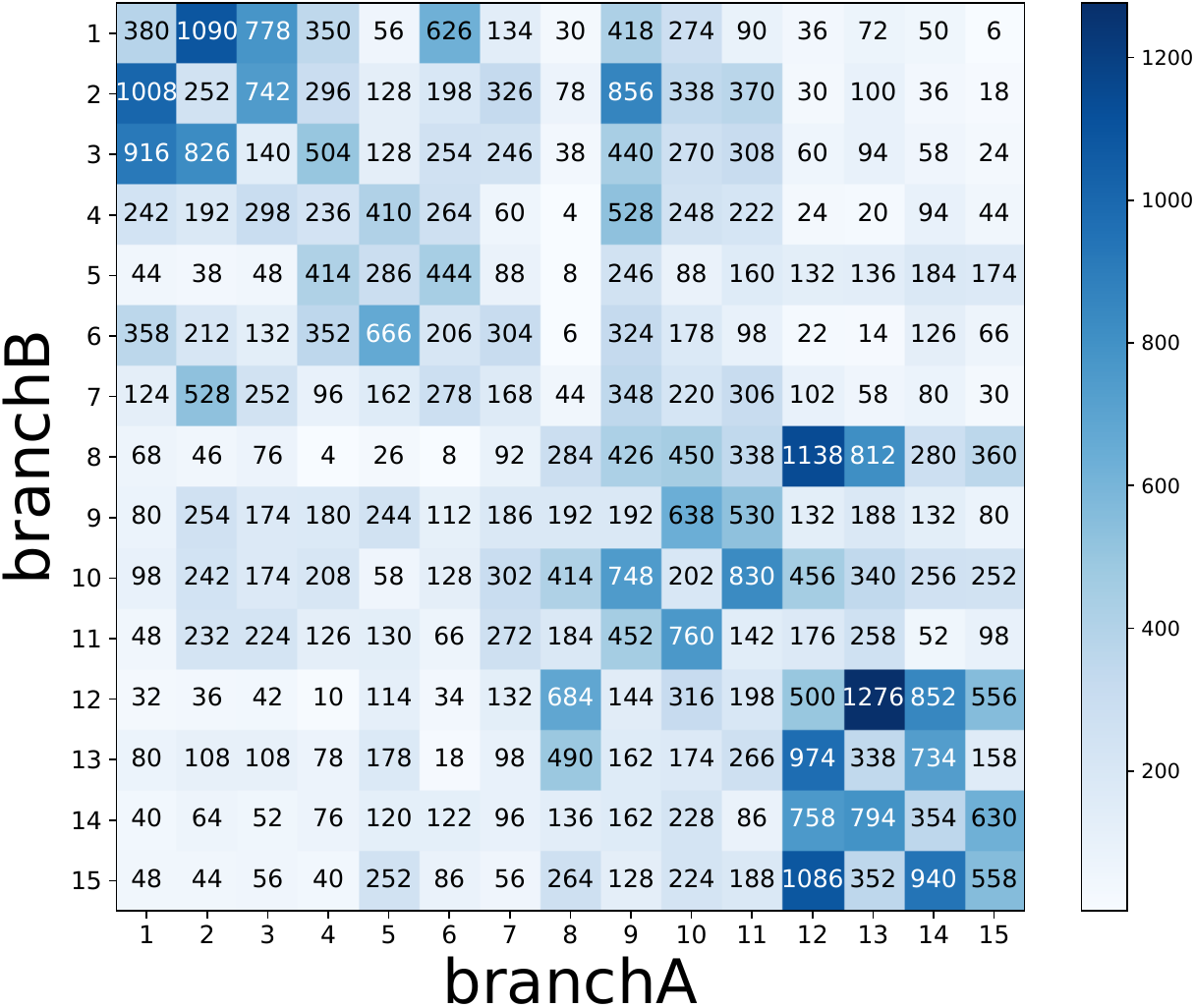}
        \label{label_for_cross_ref_2}
    }
    \vspace{-0.4cm}
    \caption{\textbf{Projection results of DB4 samples in the two branches of OPL.} We summarize the projection results of samples in the two branches and present them in the form of the confusion matrix. The projection result is based on known confidence in each branch. (a) The projection results of known samples. (b) The projection results of unknown samples.}
    \label{fig4}
\end{figure}

\textit{\textbf{Projection Inconsistency.}} Fig.~\ref{fig4} presents the projection results after the introduction of OPL.
As shown in Fig.~\ref{fig4}(a), most of the known samples are projected into the same class across both branches.
Conversely, a majority of unknown samples are projected into different classes (see Fig.~\ref{fig4}(b)).
Therefore, this projection inconsistency is effective for identifying the unknown.



\section{CONCLUSIONS}

Generalizing gesture recognition from closed set to open set is important for real-world HMI.
To tackle open set gesture recognition based on sEMG, we introduce a more effective PL method leveraging two novel and inherent distinctions between known and unknown classes.
Specifically, we propose the FAEM to strengthen the feature activation level.
Moreover, we introduce OPL to reveal the projection inconsistency of unknown samples.
Extensive experiments conducted on multiple datasets consistently demonstrate that our approach outperforms previous state-of-the-art open-set classifiers.
This means that our gesture recognition system can maintain high classification accuracy for predefined gestures while effectively rejecting gestures of disinterest. We hope this work could boost the applications of gesture recognition technologies in real-world scenarios.








\bibliographystyle{IEEEtran}
\bibliography{refs}

\end{document}